\documentclass[journal]{IEEEtran}
\usepackage{cite,graphicx,subfigure,amsmath,amssymb}
\usepackage{algorithm}
\usepackage{algorithmic}
\usepackage{multirow}
\usepackage{booktabs}
\usepackage{bm}
\usepackage{subfigure}
\usepackage[T1]{fontenc}
\usepackage{color}
\ifCLASSINFOpdf
\else
\fi
\newtheorem{Remark}{Remark}
\begin{document}

\title{A Signed Subgraph Encoding Approach via Linear Optimization for Link Sign Prediction}

\author{Zhihong~Fang, Shaolin~Tan,~\IEEEmembership{Member,~IEEE,}
        and Yaonan Wang
\thanks{%Manuscript received ??; revised ??
        Copyright (c) 2022 IEEE. Personal use of this material is permitted. However,
        permission to use this material for any other purposes must be obtained from
        the IEEE by sending an email to pubs-permissions@ieee.org. This work was supported by the National Natural Science Foundation of China under Grant 61873088.}
\thanks{Z.~Fang, S.~Tan, and Y. Wang are with the College of Electrical and Information Engineering, Hunan University, Changsha 410082, China (email: shaolintan@hnu.edu.cn).}
}%

\maketitle

\begin{abstract}

In this paper, we consider the problem of inferring the sign of a link based on limited sign data in signed networks. Regarding this link sign prediction problem, SDGNN (Signed Directed Graph Neural Networks) provides the best prediction performance currently to the best of our knowledge. In this paper, we propose a different link sign prediction architecture call SELO (Subgraph Encoding via Linear Optimization), which obtains overall leading prediction performances compared the state-of-the-art algorithm SDGNN. The proposed model utilizes a subgraph encoding approach to learn edge embeddings for signed directed networks. In particular, a signed subgraph encoding approach is introduced to embed each subgraph into a likelihood matrix instead of the adjacency matrix through a linear optimization method. Comprehensive experiments are conducted on six real-world signed networks with AUC, F1, micro-F1, and Macro-F1 as the evaluation metrics. The experiment results show that the proposed SELO model outperforms existing baseline feature-based methods and embedding-based methods on all the six real-world networks and in all the four evaluation metrics. 
 
\end{abstract}


\IEEEpeerreviewmaketitle
%%%%%%%%%%%%%%%%%%%%%%%%%%%%%%%%%%%%%%%%%%%%%%%%%%%%%%%%%%%%%%%%%%%%%%%%

%%%%%%%%%%%%%%%%%%%%%%%%%%%%%%%%%%%%%%%%%%%%%%%%%%%%%%%%%%%%%%%%%%%%%%%%
\section{Introduction}

With the proliferation of interactions on social medias, including subscribing, following, commenting, and retweeting, network structure analysis has attracted a wide range of studies. Link prediction, one main issue of network data mining, is to learn the probability of having a positive interaction between two entities based on already known positive interactions \cite{survey}. In the past several decades, the field of link prediction is eventually maturing with a wealth of well-understood methods and algorithms for mining the potential interactions \cite{SEAL,SHFF,9174790,9046288}. Nevertheless, study of link prediction only focuses on positive interactions, which is not fit for real-world networks with overt conflicts, such as distrust, opposition, disapproval \cite{kumar2018community}. A similar field, named as link sign analysis, emerges to understand the function of negative interactions in social media \cite{leskovec2010predicting}.

Similar to the link prediction problem, link sign prediction problem is to infer the sign of a target link based on the known signs of links in the network \cite{leskovec2010predicting}. Currently, there are mainly two groups of link sign prediction models: feature-based methods and network embedding methods. The feature-based methods, which are natural and widely adopted, extract some features from the graph to determine the sign of a target link. For example, All23 \cite{leskovec2010predicting} extracts a total of 23 of degree features and triad features to predict sign on the link. HOC \cite{chiang2011exploiting} generalizes the triad features of All23 to longer cycles. In \cite{beigi2020social}, three social science-guided feature groups, called EI (Emotional Information), DI (Diffusion of Innovations), and IP (Individual Personality) respectively, are comprehensive unified for signed link analysis with data sparsity. On the other hand, the network embedding methods map the given signed graph into a low-dimensional vector representation and then utilize node representation for link sign prediction \cite{BESIDE}. SGCN \cite{SGCN} utilizes the social balance theory to design a graph convolutional network (GCN) for learning the node embedding of undirected signed network. Also based on balance theory, SNEA \cite{SNEA} proposes a graph attentional layers to aggregate and propagate information from neighborhood nodes. SDGNN \cite{SDGNN} leverages four different signed directed relations to redesign attention aggregators and proposes a loss function to reconstruct signs, directions and triangles.

Nevertheless, we believe that the effectiveness of current network embedding-based methods for link sign prediction are still unsatisfactory. In fact, the present SGNN (signed graph neural network)-based methods are mainly based on pairwise interactions and triad relationships from the balance theory or status theory in signed network analysis. However, such a simplification has some limitations. First, there are many links not located in any triangles, especially for sparse networks. Second, it has shown that only less than $80\%$ triads can satisfy both the balance theory and status theory \cite{SDGNN}. Finally, if additional ``bridge'' links or $k$-group interactions were considered, better performances could be achieved \cite{BESIDE,GS-GNN}. In other words, the overall local surrounding environment, including the pairwise interaction itself and those involved triads, plays a key role to determine the sign of the link. 

Motivated by such observations, we propose an architecture called Subgraph Encoding via Linear Optimization (SELO) for link sign prediction. To embed the surrounding environment of a link, SELO consists of two modules: (1) A subgraph extractor to extract an enclosing subgraph of a target pair of nodes and (2) A subgraph encoder to encode the obtained subgraph into a vector. Such a subgraph mapping framework has been proposed in WLNM for \cite{WLNM} for unsigned link prediction. Nevertheless, the performances of link sign prediction based on the WLNM framework turn out to be unsatisfactory. The issue may come from the fact that the structural patterns for determination of the sign of a link may be quite different from those for determination of the existence of a link.

To overcome this shortcoming, we have made some key changes to the subgraph encoder in our proposed signed subgraph encoding architecture SELO. Firstly, the enclosing subgraph is represented into a likelihood matrix by a linear optimization approach instead of into a adjacency matrix. Secondly, the weights of positive links and negative links in the subgraph are assigned with different values. Finally, a novel node ordering method is proposed to refine the likelihood matrix into vectors. 

The main contributions of this paper are as follows:
\begin{enumerate}
	\item [1)] We propose a signed subgraph encoding architecture, SELO, for link sign prediction in signed graphs. In particular, a novel subgraph encoder is proposed based on a linear optimization approach, the weight-assignment strategy, and node ordering method, to encounter for the signed graph structure.
	
	\item [2)] We conduct extensive experiments on link sign prediction with real-world signed directed network datasets. The results demonstrate that the proposed SELO outperforms state-of-the-art link sign prediction methods in terms of all four evaluation metrics including AUC, F1, micro-F1, and Macro-F1 on all the real-world benchmark datasets.
\end{enumerate}

The rest paper is organized as follows. Section \ref{sec:pre} gives a brief introduction of the sign prediction problem in signed graphs and the linear optimization method. In Section \ref{seq:selo}, we introduce the framework of our proposed SELO for link sign prediction. Section \ref{seq:experiment} presents comprehensive experiments to evaluate the performance of the proposed SELO for sign prediction on a variety of real-world networks. The concluding remarks are given in Section \ref{conclusion}.

\section{Preliminaries}
\label{sec:pre}

In this section, we first present some necessary background about the sign prediction problem in signed graphs, and then give a brief introduction of the link prediction method via linear optimization, which was proposed by R. Pech \emph{et al.} in 2019 \cite{LO} and will be reformulated as a subgraph pattern encoding method in this paper.

\subsection{Signed Graph and Sign Prediction Problem}

A signed network can be formalized as $\mathcal{G}=(\mathcal{V},
\mathcal{E}^{+},\mathcal{E}^{-})$, where $\mathcal{V}=\lbrace v_{i}\rbrace _{i=1}^{N} $ is the set of nodes, and $\mathcal{E}=\mathcal{E}^{+}\cup\mathcal{E}^{-}\subseteq \mathcal{V}\times \mathcal{V}$ is the set of signed links. Here, $\mathcal{E}^{+}$ and $\mathcal{E}^{-}$ denote the set of positive links and negative links, respectively. Commonly, it is assumed that $\mathcal{E}^{+}\cap \mathcal{E}^{+}=\emptyset$. That is, each edge belongs either to the positive edge set or the negative edge set. The link sign prediction problem is about how to infer the sign type of some new or existing links with unknown sign attributes based on the observed link signs.

For analysis purpose, in the following, an adjacency matrix $A\in\mathbb{R}^{N\times N}$ is utilized to represent a signed graph, where $A_{ij}=1$ if there exists a positive link from $v_{i}$ to $v_{j}$, $A_{ij}=-1$ if there is a negative link from $v_{i}$ to $v_{j}$, and $A_{ij}=0$ if there is no link from $v_{i}$ to $v_{j}$. Furthermore, for each node $v_i\in\mathcal{V}$, we distinguish its 1-hop neighborhood as $\Gamma(v_i)= \Gamma_{out}(v_{i})\cup\Gamma_{in}(v_{i})$, where $\Gamma_{out}(v_{i})=\{v_j\in\mathcal{V}|(v_i,v_j)\in\mathcal{E}\}$,
$\Gamma_{in}(v_{i})=\{v_j\in\mathcal{V}|(v_j,v_i)\in\mathcal{E}\}$. The $2$-hop neighborhood of node $v_i$ is defined by $\Gamma^2(v_i)=\{v_j\in\mathcal{V}-\{v_i\}-\Gamma(v_i)|v_j\in \Gamma(v_k), \forall v_k\in \Gamma(v_i)\}$. In a recursive manner, the $h$-hop neighborhood of node $v_i$, denoted by $\Gamma^h(v_i)$, can be obtained. Furthermore, consider a node set $\mathcal{V}_1\subseteq\mathcal{V}$, by $\mathcal{G}(\mathcal{V}_1)$ we mean the subgraph of $\mathcal{G}$ with the node set limited to $\mathcal{V}_1$. 

\subsection{Link Prediction via Linear Optimization}

Linear optimization (LO) method for link prediction is proposed for predicting link existence in unsigned networks \cite{LO}. The method is based on a simple assumption that the likelihood of the existence of an edge between a pair of nodes is determined by the summation of contributions from their neighboring nodes. In this way, LO method transfers the link prediction problem to a problem of determining a reasonable contribution matrix to optimize the likelihood matrix.

In detail, let $S_{ij}$ denote the likelihood of the existence of a link from $v_{i}$ to $v_{j}$. The LO method assumes that the likelihood is determined by
\begin{equation}
	S_{ij} = \sum_{k} A_{ik}Z_{kj},
\end{equation}
where $Z_{kj}$ is the contribution from $v_{k}$ to $v_{j}$. The matrix form of Eq. (1) gives
\begin{displaymath}
	S = AZ.
\end{displaymath}
The LO approach realizes the link prediction objective by solving the following optimization problem
\begin{equation}
	\min_{Z}\alpha\|A-AZ\|+\|Z\|
\end{equation}
where $\|\cdot\|$ is a certain matrix norm and $\alpha$ is a free positive parameter.

Considering the Frobenius norm with power 2 (i.e. $\|Z\|=Tr(Z^TZ)$) as the matrix norm, the optimal solution of $Z$ to the above optimization problem (2) is
\begin{equation}
	Z^{*} = \alpha(\alpha A^{T}A+I)^{-1}A^{T}A \label{z}
\end{equation}
where $I$ is the identity matrix. In this case, the likelihood matrix $S$ can be obtained as
\begin{equation}
	S = AZ^{*}
\end{equation}

\begin{figure}[!t]
	\centering
	\includegraphics[width=\linewidth]{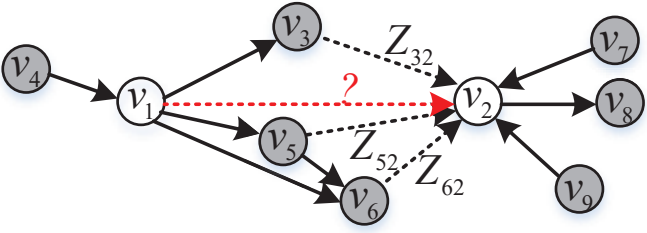}
	\caption{An illustrating example of the LO method for link prediction in directed graph. Given a graph with 11 nodes, LO first calculates contribution matrix via Eq. (\ref{z}). Then, the contributions from the outgoing neighborhood $\{v_{3},v_{5},v_{6}\}$ are summarized to determine the likelihood $S_{12}$ of the link from $v_{1}$ to $v_{2}$, i.e. $S_{12}=Z_{32}+Z_{52}+Z_{62}$. }
	\label{fig_1}
\end{figure}

Fig.\ref{fig_1} presents a simple example to illustrate how the likelihood of a target pair of nodes are determined by it enclosing subgraph based on the LO method.

%%%%%%%%%%%%%%%%%%%%%%%%%%%%%%%%%%%%%%%%%%%%%%%%%%%%%%%%%%%%%%%%%%%%%%%%%%%%%%
%%%%%%%%%%%%%%%%%%%%%%%%%%%%%%%%%%%%%%%%%%%%%%%%%%%%%%%%%%%%%%%%%%%%%%%%%%%%%%

\section{Overall Framework}
\label{seq:selo}

In this section, we propose the signed subgraph encoding architecture SELO for the sign prediction problem in signed networks. This method encodes subgraph adjacency patterns of links and feeds them into a multi-layer neural network such that the topological features from the enclosing subgraph can be automatically learned for link sign prediction. 

%%%%%%%%%%%%%%%%%%%%%%%%%%%%%%%%%%%%%%%%%%%%%%%%%%%%%%%%%%%%%%%%%%%%%%%%%%%%%%
\begin{figure*}[!t]
	\centering
	\includegraphics[width=\linewidth]{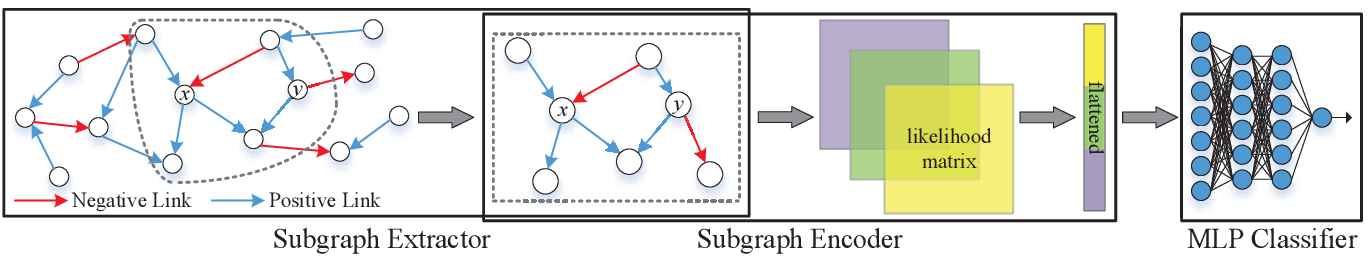}
	\caption{Illustrating 5-hop paths with diffirent topologic structure and signs.}
	\label{fig_2}
\end{figure*}
%%%%%%%%%%%%%%%%%%%%%%%%%%%%%%%%%%%%%%%%%%%%%%%%%%%%%%%%%%%%%%%%%%%%%%%%%%%%%%

The overall framework of our proposed model is depicted in Fig. 2. It consists of two modules: a subgraph extractor and a subgraph encoder.
\begin{enumerate}
	\item[i)] \textbf{Subgraph extractor:} Given a target link, the subgraph extractor extracts an enclosing subgraph from the given signed network. The extracted subgraph is deemed to contain the main topological feature of determining the sign of the link. 
	\item[ii)] \textbf{Subgraph encoder:} The subgraph encoder serves as a mapping function which turns an input subgraph topology into a fixed-size vector, which will be further input into a MLP (multi-layer perceptron) for training or prediction. 
\end{enumerate}

\subsection{Subgraph Extractor}

Subgraph extraction is the first step of subgraph embedding approaches for link-level prediction tasks. As in \cite{leskovec2010predicting}, we adopt the Breadth-First-Search (BFS) to add nodes to the subgraphs.

In detail, given a target pair of nodes $v_x, v_y\in\mathcal{V}$ or a link $(v_x,v_y)\in\mathcal{E}$, the goal is to determine an enclosing subgraph $\mathcal{G}_{xy}\subseteq\mathcal{G}$ of a fixed size $K$ from the signed graph $\mathcal{G}$. To this end, the nodes $v_x,v_y$ and their 1-hop neighbors $\Gamma(v_x), \Gamma(v_y)$ are first added to the node set $\mathcal{V}_{xy}$ of the subgraph $\mathcal{G}_{xy}$. If the number of nodes $|\mathcal{V}_{xy}|$ is less than the threshold size $K$, then the 2-hop neighbors $\Gamma^2(v_x), \Gamma^2(v_y)$ are added into the node set of subgraph. The iterative node adding process of the $h$-hop neighbors of $v_x, v_y$ ends until that $|\mathcal{V}_{xy}|\geq K$. Algorithm 1 summarizes the above subgraph extraction process.

\begin{algorithm}
	%\textsl{}\setstretch{1.8}
	\renewcommand{\algorithmicrequire}{\textbf{Input:}}
	\renewcommand{\algorithmicensure}{\textbf{Output:}}
	\caption{Subgraph extraction algorithm}
	\label{alg1}
	\begin{algorithmic}[1]
		\REQUIRE Graph $\mathcal{G}$, target node pair $(v_x,v_y)$, subgraph size $K$
		\ENSURE A enclosing subgraph $\mathcal{G}_{xy}$
		\STATE Let $\mathcal{V}_{xy}=\{v_x,v_y\}$ and $h=1$
		\STATE \textbf{while} $|\mathcal{V}_{xy}|<K$, \textbf{do}
		\STATE \qquad Let $\mathcal{V}_{xy}=\mathcal{V}_{xy}\cup \Gamma^h(v_x)\cup \Gamma^h(v_x)$
		\STATE  \qquad Let $h=h+1$
		\STATE \textbf{end while}
		\RETURN $\mathcal{G}_{xy}=\mathcal{G}(\mathcal{V}_{xy})$
	\end{algorithmic}
\end{algorithm}

Note that the returned subgraph of the above extraction algorithm may have a size larger than $K$. In this case, a number of $|\mathcal{V}(v_i,v_j)|-K$ nodes should be discarded from the subgraph. In the following, we will introduce a node ordering strategy to impose an ordering for the nodes in the subgraph. After that, the bottom $|\mathcal{V}_{xy}|-K$ nodes will be removed to make the size of all enclosing subgraphs identical to $K$.

\subsection{Subgraph Encoder}

Subgraph encoding is the core step of subgraph embedding approaches for link-level prediction tasks. Different from those subgraph encoders for link prediction in unsigned networks \cite{leskovec2010predicting}, our proposed subgraph encoder is designed for signed networks and thus several remarkable changes are made. In detail, the subgraph encoder proposed in this paper mainly contains three procedures: a link re-weighting process, a subgraph representation process, and node ordering process. We depict each of these procedures in detail in the following. 

\subsubsection{Link Re-weighting}

The goal of link re-weighting is to associate each link with a proper weight to distinguish its contribution in the subgraph. Consider a subgraph $\mathcal{G}_{xy}$ of a target pair of nodes $v_x,v_y$. Let $(v_i,v_j)$ be an arbitrary link of the subgraph. And let $d((v_{i},v_{j}),(v_{x},v_{y}))$ be the length of the shortest path to reach link $(v_{x},v_{y})$ from $(v_{i},v_{j})$ in the subgraph (Here, the direction of edges is not considered). We assign the weight of the link $(v_i,v_j)$ in the subgraph by
\begin{equation}
	\label{eq:weight}
	W_{ij}=\left\{
	\begin{aligned}
		&\frac{1}{d\left((v_{i},v_{j}),(v_{x},v_{y})\right)+1} \quad\textrm{if $(v_i,v_j)\in\mathcal{E}^+$} \\
		&\frac{-\beta}{d\left((v_{i},v_{j}),(v_{x},v_{y})\right)+1} \quad\textrm{otherwise $(v_i,v_j)\in\mathcal{E}^-$}
	\end{aligned}	
	\right.
\end{equation}
Here, $\beta$ is a positive scaling factor.

The above link weighting strategy generates a hierarchical structure of the subgraph according to the distance to target nodes. It is implicitly assumed that the importance of a link in subgraph to the target link is inversely proportional to the distance between them. Such a hierarchical structure has been utilized to promote the link prediction performance in WLNM \cite{WLNM}. Compared to the non-weighted adjacency matrix, it was shown that the weighted matrix gave a better reflection of how the subgraph pattern for link level prediction tasks.

Furthermore, in our proposed weighting strategy Eqs. (\ref{eq:weight}), we assign different weight to the positive and negative links. The motivation is intuitive as follows. On the one hand, the negative information is perceived to be more noticeable and more credible. In fact, it has been reported that behavioral decision is more easily affected by negative information than positive information \cite{cho2006mechanism}. On the other hand, negative links provide a significant amount of additional complementary knowledge for positive links \cite{tang2016survey,tang2014distrust}. Experimental results have shown that a small number of negative links can improve recommendation performance \cite{ma2009learning} in link prediction. For sign prediction problem, features extracted from negative edges significantly improves the prediction accuracy \cite{guha2004propagation,leskovec2010predicting}.

In the following, the scaling factor will be assigned to be 
\begin{equation}
	\beta= 1+\log_{10}\frac{n^{+}}{n^{-}},
\end{equation}
as a benchmark. Here $n^{+}=|\mathcal{E}^+|$ and $n^{-}=|\mathcal{E}^-|$ are the number of positive links and negative links in the signed network, respectively.
As positive links commonly outnumber the negative links by a huge margin in signed graphs, we can find that the benchmark $\beta>1$. In the experimental section, we also consider other choice of $\beta$. It will be shown that such a re-weight strategy indeed promotes the performance of link sign prediction.

\subsubsection{Subgraph Representation}

The goal of subgraph representation is to encode the subgraph structure into a readable data for subsequent neural network training or prediction. A simple and direct approach is to introduce a consistent node ordering algorithm for labeling the nodes and then generate a corresponding adjacency matrix of the subgraph as the input data. In \cite{LO}, the Weisfeiler-Lehman algorithm was adapted to rank the nodes. In \cite{Devi}, the geometric mean distance and arithmetic mean distance were utilized as the ordering measure for labeling nodes. These node ordering methods are efficient for node labeling in unsigned networks. Yet, the sign of edges is not considered and thus will be ignored when utilizing these methods in signed networks.

In this paper, we utilize the likelihood matrix obtained from the LO method, instead of adjacency matrix, to encode the subgraph pattern of the signed subgraph. In detail, given an enclosing subgraph $\mathcal{G}_{xy}$ of a target pair of nodes $v_x,v_y$. We relabel the target node pair $v_x,v_y$ as node 1 and 2, respectively. All the other nodes in the subgraph are randomly labeled from 3 to $|\mathcal{V}_{xy}|$ initially. In this case, an adjacency matrix $A$ of the subgraph can be obtained. From the link re-weighting in Eqs. (\ref{eq:weight}), we can further obtain a weight matrix $W$ of the subgraph.

Following the LO method, we derive the likelihood matrix $S^{(1)}$ of the subgraph by
\begin{equation}
	S^{(1)}=WZ^{(1)},
\end{equation}
where the contribution matrix $Z^{(1)}$ is determined by
\begin{equation}
	\label{opt1}
	\min_{Z^{(1)}}\alpha\|W-WZ^{(1)}\|+\|Z^{(1)}\|.
\end{equation}
Here, $\alpha$ is a free positive parameter. Considering the Frobenius norm with power 2 as the matrix norm, the solution to the likelihood matrix gives
\begin{equation}
	\label{eq:s1}
	S^{(1)}=\alpha W(\alpha W^{T}W+I)^{-1}W^{T}W.
\end{equation}

Following a similar argument of LO method, we can also derive the likelihood matrix $S^{(2)}, S^{(3)}, S^{(4)}$ of the subgraph by
\begin{equation}
	\left\{
	\begin{aligned}
	&S^{(2)}=W^TZ^{(2)} \\
	&S^{(3)}=Z^{(3)}W^{T} \\
	&S^{(4)}=Z^{(4)}W
	\end{aligned}
    \right.
\end{equation}
where the contribution matrix $Z^{(2)}, Z^{(3)}$, and $Z^{(3)}$ are determined by
\begin{equation}
	\left\{
	\begin{aligned}
	&\min_{Z^{(2)}}\alpha\|W-W^{T}Z^{(2)}\|+\|Z^{(2)}\|\\
	&\min_{Z^{(3)}}\alpha\|W-Z^{(3)}W^{T}\|+\|Z^{(3)}\|\\
	&\min_{Z^{(4)}}\alpha\|W-Z^{(4)}W\|+\|Z^{(4)}\|
    \end{aligned}
    \right.
\end{equation}
Considering the Frobenius norm with power 2 as the matrix norm, the solutions of the above likelihood matrices are
\begin{equation}
	\label{eq:likelihood_2}
	\left\{
	\begin{aligned}
	&S^{(2)}=\alpha W^{T}(\alpha WW^{T}+I)^{-1}W^2\\
	&S^{(3)}=\alpha W^2(\alpha W^{T}W+I)^{-1}W^{T}\\
	&S^{(4)}=\alpha WW^{T}(\alpha WW^{T}+I)^{-1}W.
	\end{aligned}
     \right.
\end{equation}

Overall, given a subgraph $\mathcal{G}_{xy}$ with a weight matrix $W$, we can get four likelihood matrices $S^{(1)}~S^{(4)}$. Nevertheless, it can be easily checked that $S^{(1)}=S^{(4)}$. Thus, the final subgraph representation in this paper is $(S^{(1)},S^{(2)},S^{(3)})$.

\begin{Remark}[Computation complexity of subgraph representation]
	The computation complexity of the proposed subgraph representation is very low, though it involves quite a lot matrix manipulation such as inversion operation. The underlying reason are of two-fold. First, the size of the subgraph $K$ is very small compared to the size of the entire network. Thus, the computation complexity is irrelevant with the size of the whole network. Second, the parameter is often selected to be closely to 0. In this case, considering the Neumann series of these likelihood matrices, we can get  
	\begin{equation}
		\label{eq:expansion}
		\left\{
		\begin{aligned}
			S^{(1)}=\sum_{k=0}^{\infty}(-1)^{k}\alpha^{k+1}W(W^{T}W)^{k+1},\\
			S^{(2)}=\sum_{k=0}^{\infty}(-1)^{k}\alpha^{k+1}(W^{T}W)^{k+1}W,\\
			S^{(3)}=\sum_{k=0}^{\infty}(-1)^{k}\alpha^{k+1}W(WW^{T})^{k+1}.
		\end{aligned}
        \right.
	\end{equation} 
   In this case, we can approximate the likelihood matrix by the first one or two items in the Neumann series. Hence, the matrix inverse operation is not required.
\end{Remark}

%%%%%%%%%%%%%%%%%%%%%%%%%%%%%%%%%%%%%%%%%%%%%%%%%%%%%%%%%%%%%%%%%%%%%%%%%%%%%%
\begin{figure}[!t]
	\centering
	\includegraphics[width=\linewidth]{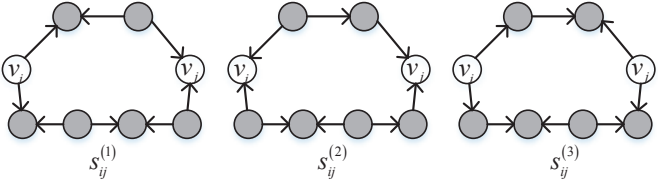}
	\caption{Illustration of the 3-hop and 5-hop paths of $S^{(1)}$, $S^{(2)}$ and $S^{(3)}$, respectively.}
	\label{fig_3}
\end{figure}
%%%%%%%%%%%%%%%%%%%%%%%%%%%%%%%%%%%%%%%%%%%%%%%%%%%%%%%%%%%%%%%%%%%%%%%%%%%%%%

%%%%%%%%%%%%%%%%%%%%%%%%%%%%%%%%%%%%%%%%%%%%%%%%%%%%%%%%%%%%%%%%%%%%%%%%%%%%%%
\begin{figure*}[!t]
	\centering
	\includegraphics[width=\linewidth]{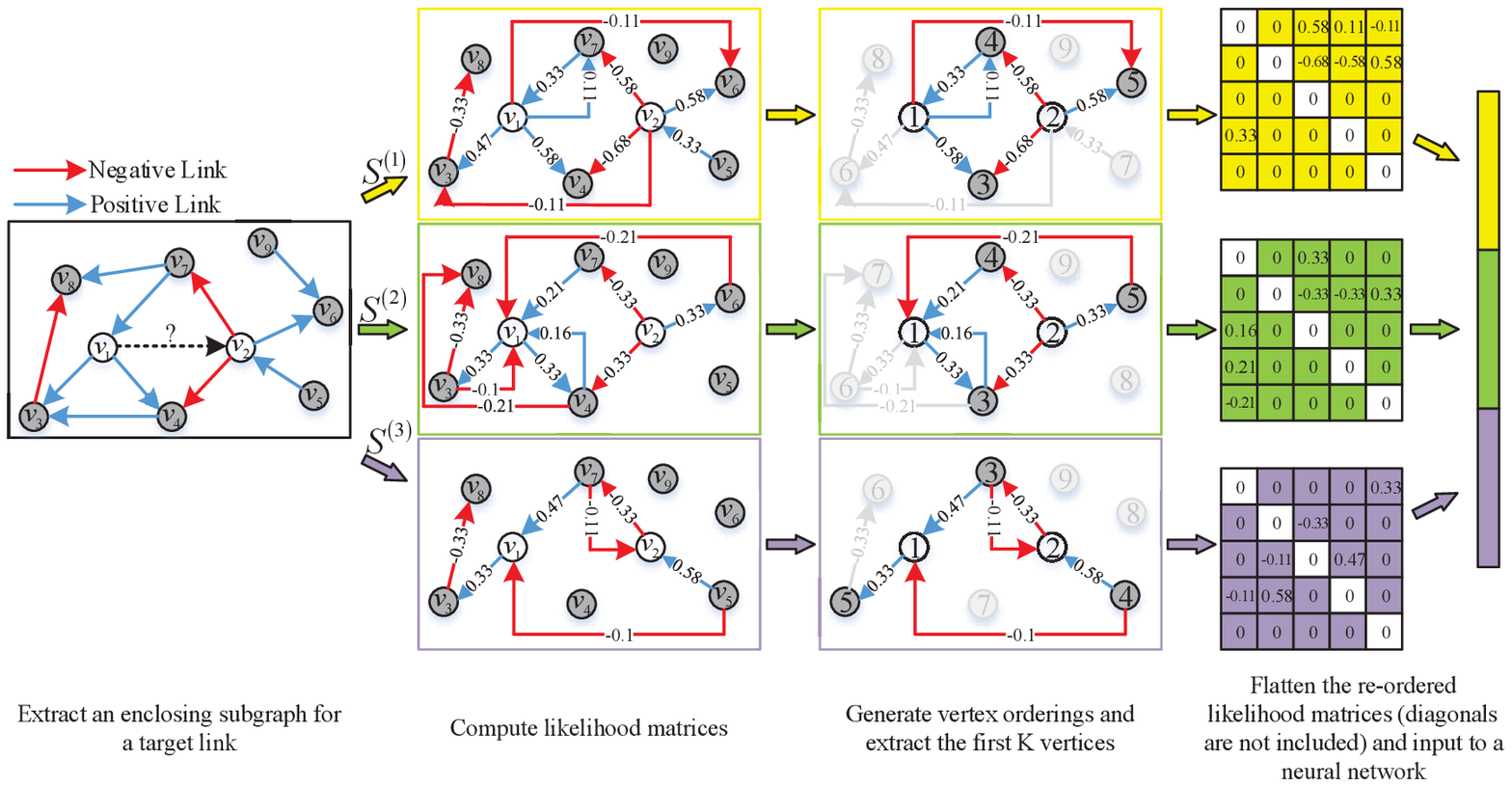}
	\caption{Illustration of the subgraph encoding process. The subgraph encoding processes by the likelihood matrices $S^{(1)}$, $S^{(2)}$ and $S^{(3)}$ are distinguished with different colors. Finally, the three likelihood matrices are concatenated and fed into a neural network for link sign prediction.}
	\label{fig_4}
\end{figure*}
%%%%%%%%%%%%%%%%%%%%%%%%%%%%%%%%%%%%%%%%%%%%%%%%%%%%%%%%%%%%%%%%%%%%%%%%%%%%%%

\begin{Remark}[Topological interpretation of the likelihood matrices]
	From the Neumann series of these likelihood matrices, it can be found the elements in these matrices are determined by the odd directed paths in the subgraph. As shown in Fig. \ref{fig_3}, the $(i,j)$-entry of $S^{(1)}$ is summation of weigh product for all 3-hop paths of the form $v_{i}\to v_{x}\gets v_{y}\to v_{j}$ in the subgraph. Similarly, the likelihood matrix $S^{(2)}$ and $S^{(3)}$ are mainly determined by paths of the form $v_{i}\gets v_{x} \to v_{y}\to v_{j}$ and $v_{i} \to v_{x} \to v_{y} \gets v_{j}$, respectively. Thus, it can be observed that the embedded topological features in the likelihood matrix are complementary to each other.
\end{Remark}

\subsubsection{Subgraph Node Ordering}

The goal of subgraph node ordering is to prune the subgraph to fixed size and to re-label its nodes with a consistent order. Indeed, note that the size of the resulted likelihood matrices $(S^{(1)},S^{(2)},S^{(3)})$ may be varying for different subgraphs, since the returned subgraph by the extraction algorithm may have a size larger than $K$. Furthermore, the entries in the likelihood matrix may be not fixed even for the same subgraph, due to the randomly labeling of the nodes in the subgraph. To facilitate the training of neural network, a consistent node ordering method is required. In unsigned network representation methods, the Weisfeiler-Lehman algorithm is commonly adopted for graph labelling \cite{WL}. In this paper, we evaluate the importance of nodes in the subgraph based on the likelihood matrix obtained by the above subgraph representation process. 

In detail, given a likelihood matrix $S^{(k)}$ with $k=1,2,3$ of an enclosing subgraph $\mathcal{G}_{xy}$, we can associate an importance score to all the nodes $v_i\in \mathcal{G}_{xy}, v_i\neq v_x,v_i\neq v_y$ by:
\begin{equation}
	\label{eq:importance score}
	b^{(k)}_{i}=|s^{(k)}_{xi}|+|s^{(k)}_{ix}|+|s^{(k)}_{yi}|+|s^{(k)}_{iy}|,
\end{equation}
where $s^{(k)}_{ij}$ is the $ij$-th entry of the likelihood matrix $S^{(k)}$. Note that the above measurement $b^{(k)}_{i}$ is indeed a summation of the likelihood of existing an out-edge and in-edge between node pairs $v_x, v_i$ and $v_y, v_i$.

Given the above importance score, we can relabel the nodes as follows: 1) the target nodes $v_x$ and $v_y$ are labeled as node 1 and 2; 2) all the other nodes in the subgraph is labeled from 3 to $|\mathcal{V}_{xy}|$ in a descending manner according the importance score. In this case, all the three likelihood matrices $S^{(k)}$ with $k=1,2,3$ can be rearranged by this consistent node labeling. Furthermore, the size of the likelihood matrix is constrained to a fixed size $K$ by removing the bottom $|\mathcal{V}_{xy}|-K$ nodes from the subgraph.

\subsubsection{The Overall Subgraph Encoding Algorithm}

The overall framework of subgraph encoding process is illustrated in Fig.\ref{fig_4}. In detail, given a subgraph of a target pair of nodes, firstly a weighted matrix is obtained by the link re-weighting procedure. Then, the likelihood matrices $S^{(1)}$, $S^{(2)}$, $S^{(3)}$ are computed by Eqs. (\ref{eq:s1}) and (\ref{eq:likelihood_2}). In the last, each likelihood matrix $S^{(k)}$ will be re-ordered according to the descending order of its corresponding importance score with a fixed size. The resulted $S^{(1)}$, $S^{(2)}$, $S^{(3)}$ are flattened and concatenated for training the neural networks. The detailed subgraph encoding algorithm is presented in Algorithm 2.

\begin{algorithm}
	%\textsl{}\setstretch{1.8}
	\renewcommand{\algorithmicrequire}{\textbf{Input:}}
	\renewcommand{\algorithmicensure}{\textbf{Output:}}
	\caption{Subgraph encoding algorithm}
	\label{alg1}
	\begin{algorithmic}[1]
		\REQUIRE A subgraph $\mathcal{G}_{xy}$ of a target pair of nodes $v_{x}$, $v_{y}$
		\ENSURE A likelihood matrix $S^{(1)},S^{(2)},S^{(3)}$ of size $K$
		\STATE Assign initial label: Label node $v_{x}$, $v_{y}$ as node 1 and 2; randomly label the other nodes in the subgraph as node 3 to $|\mathcal{V}_{xy}|$
		\STATE Get the adjacency matrix and translate it into the weight matrix by the re-weighting strategy (\ref{eq:weight})
		\STATE Calculate the likelihood matrix $S^{(1)}$ by Eq. (\ref{eq:s1}) and $S^{(2)},S^{(3)}$ by Eq. (\ref{eq:likelihood_2})
		\STATE \textbf{For} $k=1,2,3$, \textbf{Do}
		\STATE \qquad Calculate the importance score $b^{(k)}_{i}$ by Eq. (\ref{eq:importance score}) for all nodes $i=3,\cdots,|\mathcal{V}_{xy}|$
		\STATE \qquad Refine the likelihood matrix $S^{(k)}$ by relabeling the nodes in the descending order and discarding the bottom $|\mathcal{V}_{xy}|-K$ nodes
		\RETURN $S^{(1)},S^{(2)},S^{(3)}$
	\end{algorithmic}
\end{algorithm}

\subsection{Neural Network Training and Testing}

After encoding the enclosing subgraphs, the next step is to train a binary classifier for link sign prediction. Considering its powerful representation capability, the fully connected neural network is employed in this paper to learn the complicated structure features of the subgraph. In detail, given a signed graph $\mathcal{G}=(\mathcal{V},\mathcal{E}^{+},\mathcal{E}^{-})$, we randomly sample $80\%$ links in $\mathcal{E}^{+}$ as the positive samples, and $80\%$ links in $\mathcal{E}^{-}$ as the negative samples. For a given link $(v_x,v_y)$ in the training set, we extract its enclosing subgraph and encode it into the likelihood matrices using the above signed subgraph encoding architecture. These likelihood matrices are then flattened and input into the neural networks with their labels ($1:(v_x,v_y)\in \mathcal{E}^{+}, -1:(v_x,v_y)\in \mathcal{E}^{-}$).

After training the neural network, we can use it as a classifier to predict the sign of a testing link. The procedure is similar: an enclosing subgraph of the testing link is extracted first, then the likelihood matrices are determined by the subgraph encoder, and lastly the flattened and concatenated matrices are feed into the pre-trained neural network. The neural network will output a score between 0 and 1 for each testing link, which indicates the probability for the link to be positive or not.

\section{Experimental Results and Analysis}

In this section, we conduct comprehensive experiments to evaluate the performance of our proposed SELO for link sign prediction on a variety of real-world networks. The results shows that our SELO gives the state-of-the-art performance in various evaluation metrics.

\subsection{Datasets and Evaluation Metrics}

Five real-world networks, including Bitcoin-Alpha, Bitcoin-OTC, WikiRfA, Slashdot and Epinions, are adopted as the experimental datasets in this paper. Bitcoin-Alpha and Bitcoin-OTC are two who-trusts-whom networks of users, who trade with Bitcoin on the platform called Bitcoin Alpha and Bitcoin OTC, respectively. WikiRfA is signed voting networks of Wikipedia users, where each user can either support (positive link) or oppose (negative link) the promotion of other users by voting. Slashdot is a technology-related news website, which introducs Slashdot Zoo to allow users to tag each other as friends or foes. Epinions is a general consumer review site, which builds the who-trust-whom online social network of consumers based their trust or distrust to others' reviews. Table \ref{tab:dadaset} comprises some key statistical information of each network.

The area under curve (AUC), F1 score, macro-F1 and micro-F1 are utilized as the standard metrics to evaluate the link sign prediction performance. Each dataset is randomly split into a training set and a testing set for 5 times. The average value of the above four performance metrics will be recorded for comparison.

%%%%%%%%%%%%%%%%%%%%%%%%%%%%%%%%%%%%%%%%%%%%%%%%%%%%%%%%%%%%%%%%%%%%%%%%%%%%%%
\begin{table}
	\renewcommand{\arraystretch}{1.3}
	\caption{Some statistical information of each real-world signed networks. Nodes, Pos links, Neg links, Ratio refer to the number of nodes, number of positive links, number of negative links and the ratio of Positive links to Negative links, respectively.}
	\label{tab:dadaset}
	\centering
	\begin{tabular}{c|cccc}
		\hline
		Data &Nodes&Pos links&Neg links&Ratio\\
		\hline
		Bitcoin-Alpha&3782&22649 &1536 &14.75\\
		Bitcoin-OTC&5881&32028 &3563 &8.99\\
		WikiRfA&11259&138813 &39283 &3.53\\
		Slashdot&82140&425071 &124130 &3.42\\
		Epinions&131827&717667 &123704 &5.80\\
		\hline
	\end{tabular}
\end{table}
%%%%%%%%%%%%%%%%%%%%%%%%%%%%%%%%%%%%%%%%%%%%%%%%%%%%%%%%%%%%%%%%%%%%%%%%%%%%%%

\subsection{Baselines}

To evaluate the performance of the proposed SELO, we compare it with the following baseline methods for link sign prediction in all the above five real-world datasets.

%%%%%%%%%%%%%%%%%%%%%%%%%%%%%%%%%%%%%%%%%%%%%%%%%%%%%%%%%%%%%%%%%%%%%%%%%%%%%%
\begin{table*}[tbp]
	\begin{center}
		\renewcommand{\arraystretch}{1.3}
		\centering
		\caption{Comparison with other sign prediction algorithms.}
		\label{tab:result}
		\resizebox{\textwidth}{!}{
			\begin{tabular}{c|c|c|ccc|cccccc|c}
				\toprule
				%				\multirow{2}{*}{Dataset}&\multirow{2}{*}{Metric}&Feature Engineering&\multicolumn{3}{c}{Unsigned Network Embedding} &\multicolumn{6}{|c|}{Signed Network Embedding}&Subgraph Pattern\\
				%				&&All23&Deepwalk&Node2vec&LINE&SiNE&SIGNet&BESIDE&SGCN&SiGAT&SDGNN&SELO\\
				
				Dataset&Metric&All23&Deepwalk&Node2vec&LINE&SiNE&SIGNet&BESIDE&SGCN&SiGAT&SDGNN&SELO\\
				\midrule
				\multirow{4}{*}{Bitcoin-Alpha}&Micro-F1&0.9486&0.9367&0.9355&0.9352&0.9458&0.9422&0.9489&0.9256&0.9456&0.9491&\textbf{0.9599}\\
				&F1&0.9730&0.9673&0.9663&0.9664&0.9716&0.9696&0.9732&0.9607&0.9714&0.9729&\textbf{0.9788}\\
				&Macro-F1&0.7167&0.4848&0.6004&0.5220&0.6869&0.6965&0.7300&0.6367&0.7026&0.7390&\textbf{0.8107}\\
				&AUC&0.8882&0.6409&0.7576&0.7114&0.8728&0.8908&0.8981&0.8469&0.8872&0.8988&\textbf{0.9187}\\
				\hline
				\multirow{4}{*}{Bitcoin-OTC}&Micro-F1&0.9361&0.8937&0.9089&0.8911&0.9095&0.9229&0.9320&0.9078&0.9268&0.9357&\textbf{0.9553}\\
				&F1&0.9653&0.9434&0.9507&0.9413&0.9510&0.9581&0.9628&0.9491&0.9602&0.9647&\textbf{0.9754}\\
				&Macro-F1&0.7826&0.5281&0.6793&0.5968&0.6805&0.7386&0.7843&0.7306&0.7533&0.8017&\textbf{0.8656}\\
				&AUC&0.9121&0.6596&0.7643&0.7248&0.8571&0.8935&0.9152&0.8755&0.9055&0.9124&\textbf{0.9532}\\
				\hline
				\multirow{4}{*}{WikiRfA}&Micro-F1&0.8346&0.7837&0.7814&0.7977&0.8338&0.8384&0.8589&0.8489&0.8457&0.8627&\textbf{0.8644}\\
				&F1&0.8987&0.8779&0.8719&0.8827&0.8972&0.9001&0.9117&0.9069&0.9042&0.9142&\textbf{0.9155}\\
				&Macro-F1&0.7235&0.4666&0.5626&0.5738&0.7319&0.7384&0.7803&0.7527&0.7535&0.7849&\textbf{0.7859}\\
				&AUC&0.8604&0.5876&0.6930&0.6772&0.8602&0.8682&0.8981&0.8563&0.8829&0.8898&\textbf{0.9049}\\
				\hline
				\multirow{4}{*}{Slashdot}&Micro-F1&0.8472&0.7738&0.7526&0.7489&0.8265&0.8389&0.8590&0.8296&0.8494&0.8616&\textbf{0.8818}\\
				&F1&0.9070&0.8724&0.8528&0.8525&0.8918&0.8983&0.9105&0.8926&0.9055&0.9128&\textbf{0.9255}\\
				&Macro-F1&0.7399&0.4384&0.5390&0.5052&0.7273&0.7554&0.7892&0.7403&0.7671&0.7892&\textbf{0.8198}\\
				&AUC&0.8880&0.5408&0.6709&0.6145&0.8409&0.8752&0.9017&0.8534&0.8874&0.8977&\textbf{0.9250}\\
				\hline
				\multirow{4}{*}{Epinions}&Micro-F1&0.9226&0.8214&0.8563&0.8535&0.9173&0.9113&0.9336&0.9112&0.9293&0.9355&\textbf{0.9556}\\
				&F1&0.9561&0.9005&0.9170&0.9175&0.9525&0.9489&0.9615&0.9486&0.9593&0.9628&\textbf{0.9743}\\
				&Macro-F1&0.8130&0.5131&0.6862&0.6305&0.8160&0.8060&0.8601&0.8105&0.8454&0.8610&\textbf{0.9055}\\
				&AUC&0.9444&0.6702&0.8081&0.6835&0.8872&0.9095&0.9351&0.8745&0.9333&0.9411&\textbf{0.9722}\\
				\bottomrule
		\end{tabular}}
	\end{center}
\end{table*}
%%%%%%%%%%%%%%%%%%%%%%%%%%%%%%%%%%%%%%%%%%%%%%%%%%%%%%%%%%%%%%%%%%%%%%%%%%%%%%

\begin{enumerate}
	\item[1)] Feature Engineering: All23 \cite{leskovec2010predicting} extracts a total of 23 features from 1-hop neighborhood of the target link based on the sociological balance theory and status theory. The first class of features focus on the degree information of a given node pair $v_{i}$ and $v_{j}$. The second class of features are different types of the triads that contains $v_{i}$ and $v_{j}$. The 23 features are finally fed into a logistic regression model.
	\item[2)] Unsigned Network Embedding: We provide three classical unsigned network embedding methods to link sign prediction task, which contain DeepWalk \cite{deepwalk}, Node2vec \cite{node2vec} and LINE \cite{LINE}. Owing to these methods are designed for unsigned network, we remove the negative links in the training stage.
	\item[3)] Signed Network Embedding: There existing a good number of signed network embedding methods. We choose six signed network embedding methods, including SiNE \cite{SiNE}, SIGNet \cite{SIGNet}, BESIDE \cite{BESIDE}, SGCN \cite{SGCN}, SiGAT \cite{SiGAT}, SDGNN \cite{SDGNN}.	
\end{enumerate}

\subsection{Prediction Performance}

\label{seq:experiment}

Table \ref{tab:result} presents the average Micro-F1, F1, Macro-F1, and AUC of link sign prediction on different datasets. The hyper-parameters are as follows. For the proposed SELO, we set the subgraph size $K=5$ and the free parameters $\alpha=0.005$. For the neural network structure, we use a fully-connected neural network with 3 hidden layers of 32, 32, 16 neurons respectively and a softmax output layer. We adopt the Rectified linear unit (ReLU) as the activation function for all hidden layers and the Adam update rule for optimization \cite{adam} with mini-batch size 512 and learning rate 0.001. The number of epochs is set to 100 for all the six datasets. The hyper-parameters in baseline methods are set as suggested by the \cite{SDGNN}.

From Table \ref{tab:result}, it can be observed that our proposed SELO gives the best performances on all the six real-world datasets and in all terms of evaluation metrics. In particular, an apparent performance improvement can be observed for SELO on Bitcoin-Alpha, Bitcoin-OTC and Slashdot networks. Furthermore, the proposed SELO outperforms other baseline methods by a large margin in terms of Macro-F1 and AUC. These results demonstrate the capability of our proposed SELO in link sign prediction problems.

\subsection{Time Complexity}

In this subsection, we analyze the time complexity of the framework SELO. For simplicity, let $\overline{n}$ denote the average nodes number in extracted subgraphs. The link re-weighting procedure assigns a weight for each link based on the distance between each link with the target link, which results a computation complexity of $O(\overline{n}^{3})$. The subgraph representation procedure contains matrix inverse and multiplication operations, which has a computational complexity of $O(\overline{n}^{3})$. Furthermore, the subgraph node ordering procedure is decided by a sorting algorithm with time complexity $O(\overline{n}log\overline{n})$. Overall all, the time complexity of the signed subgraph encoding process is $O(\overline{n}^{3})+O(\overline{n}log\overline{n})\approx O(\overline{n}^{3})$. We adopt MLP as classfier, whose time complexity is negligible compared to the subgraph encoding process. The total time complexity of SELO could be approximated to $O(|\mathcal{E}|\overline{n}^{3})$. 

Table \ref{tab:time} gives a further comparison of the required computation time between SELO and SDGNN. SELO first encodes subgraphs to features, then fed these features into a MLP. SDGNN leverages a GNN framework to obtain node embedding, then the embeddings of target nodes are concatenated as features and then fed into a logistic regression model. Therefore, we record subgraph encoding time and MLP training time for SELO and record the training time of GNN model and logistic regression model for SDGNN, respectively. The experiments in Table \ref{tab:time} run on a computer with 11th Gen Intel(R) Core(TM) i5-11260H CPU and the code of SELO and SDGNN is written in Python 3.8. We conduct experiments with different train-test splits for 5 times to get the average computation time and its standard deviation. It can be observed that the time required for SELO is significantly less than that for SDGNN.

%%%%%%%%%%%%%%%%%%%%%%%%%%%%%%%%%%%%%%%%%%%%%%%%%%%%%%%%%%%%%%%%%%%%%%%%%%%%%%
	\begin{table*}[tbp]
		\begin{center}
			\renewcommand{\arraystretch}{1.2}
			\centering
			\caption{Comparison of computation time (Units are seconds).}
			\label{tab:time}
			\begin{tabular}{c|ccc|ccc}
				\toprule
				\multirow{2}{*}{Dataset}&\multicolumn{3}{c}{SELO}&\multicolumn{3}{c}{SDGNN}\\
				&Subgraph Pattern Encoding&MLP Training&Total&GNN Training&Ligistic Regression Training&Total\\
				\midrule
				Bitcoin-Alpha&78.38$\pm$0.60 &3.36$\pm$0.04 &81.74$\pm$0.64 &774.67$\pm$11.92&0.10$\pm$0.01&774.77$\pm$11.91\\
				Bitcoin-OTC&248.84$\pm$2.25 &4.96$\pm$0.14 &253.80$\pm$2.23 &1848.51$\pm$8.61&0.13$\pm$0.01&1848.63$\pm$8.62\\
				\bottomrule
			\end{tabular}
		\end{center}
	\end{table*}
%%%%%%%%%%%%%%%%%%%%%%%%%%%%%%%%%%%%%%%%%%%%%%%%%%%%%%%%%%%%%%%%%%%%%%%%%%%%%%

\subsection{Hyper-parameter Analysis}

In this subsection, we analyze the effect of hyper-parameters on the performances of SELO. In detail, there are mainly two hyper-parameters to encode the subgraph patterns in the proposed SELO. The first parameter is the positive scaling factor $\beta$ in the re-weighting strategy (\ref{eq:weight}) for assigning a weight to negative links. The second parameter is the positive coefficient $\alpha$ for determining the likelihood matrices. In this following, we investigate how the SELO performs with different setting of these hyper-parameters by simulations. Due to space limit, we only present the experimental results about hyper-parameter analysis on the Bitcoin-Alpha dataset. Similar observations are preserved for other real-world datasets.

%%%%%%%%%%%%%%%%%%%%%%%%%%%%%%%%%%%%%%%%%%%%%%%%%%%%%%%%%%%%%%%%%%%%%%%%%%%%%%
\begin{figure}[!t]
	\centering
	\includegraphics[width=\linewidth]{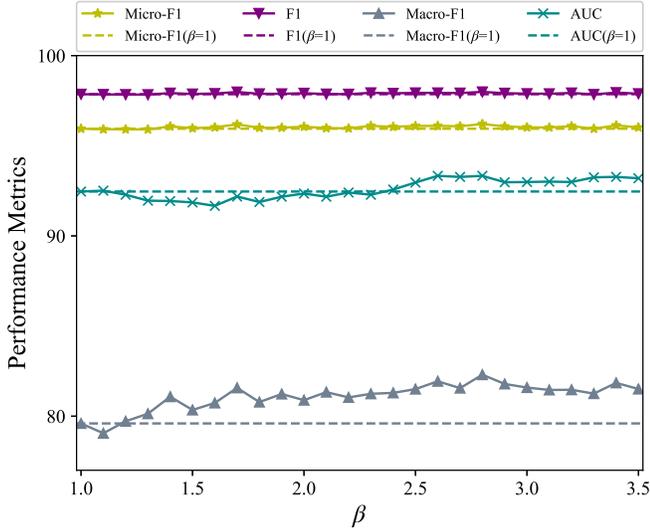}
	\caption{The effect of negative link weight scaling factor $\beta$ on the link sign prediction performances. Experiment is conducted in Bitcoin-Alpha dataset.}
	\label{fig_b}
\end{figure}
%%%%%%%%%%%%%%%%%%%%%%%%%%%%%%%%%%%%%%%%%%%%%%%%%%%%%%%%%%%%%%%%%%%%%%%%%%%%%%

Fig. \ref{fig_b} records the Micro-F1 and AUC performances by varying $\beta$ from $1$ to $3.5$. It can be observed that the Micro-F1 and AUC performances are obviously increased by assigning $\beta>1$ instead of $\beta=1$. In other words, it could promote the link sign prediction performances of SELO by assigning a higher weight to the negative links than that of positive links. Furthermore, when $\beta\geq1+\log_{10}(n^+/n^-)\approx 2$, the correlation between Micro-F1 and AUC performances and $\beta$ is not significant. That is why the scaling factor $\beta$ is assign by Eq. (6) as a benchmark.

\begin{figure}[!t]
	\centering
	\includegraphics[width=\linewidth]{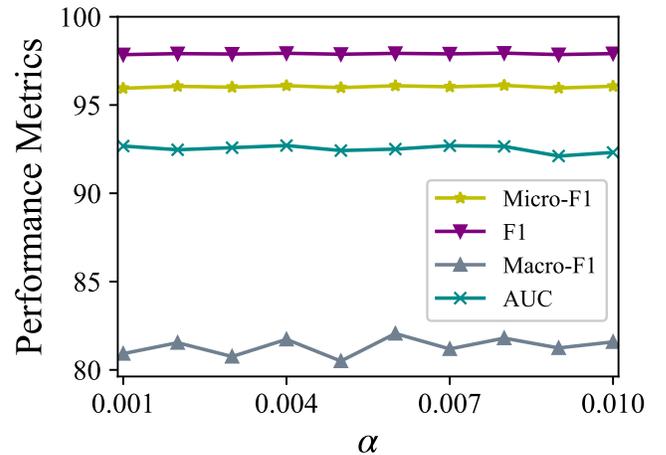}
	\caption{Robustness analysis of the prediction performances with varying $\alpha$.}
	\label{fig_a}
\end{figure}

Fig. \ref{fig_a} depicts the prediction performances by varying $\alpha$ from $0.001$ to $0.01$. From Eqs. (\ref{eq:s1}) and (\ref{eq:likelihood_2}), it can be found that the free parameter $\alpha$ determine the contributions of odd paths with different hops. The contribution of large-hop paths to the likelihood matrix can be ignored when the $\alpha$ is enough small. Nevertheless, from Fig. \ref{fig_a} it can be observed that all terms of metrics are nearly stationary to change of these parameters. Hence, the performance of SELO is not sensitive to choice of the parameter $\alpha$.

%%%%%%%%%%%%%%%%%%%%%%%%%%%%%%%%%%%%%%%%%%%%%%%%%%%%%%%%%%%%%%%%%%%%%%%%%%%%%%
%%%%%%%%%%%%%%%%%%%%%%%%%%%%%%%%%%%%%%%%%%%%%%%%%%%%%%%%%%%%%%%%%%%%%%%%%%%%%%

%%%%%%%%%%%%%%%%%%%%%%%%%%%%%%%%%%%%%%%%%%%%%%%%%%%%%%%%%%%%%%%%%%%%%%%%%%%%%%%
%\begin{table}[tbp]
%	\begin{center}
%		\renewcommand{\arraystretch}{1.2}
%		\centering
%		\caption{Illustrating the comparsion of encoding sugraph by only $S^{(1)}$, $S^{(1)}$, $S^{(1)}$ and their concated.}
%		\label{tab:single_model}
%		\begin{tabular}{cc|cccc}
%			\toprule
%			Dataset&Method&Micro-F1&F1&Macro-F1&AUC\\
%			\midrule
%			\multirow{6}{*}{Bitcoin-Alpha}&$A$&0.9479&0.9726&0.7260&0.8823\\
%			&$W$&0.9466&0.9717&0.7424&0.9170\\
%			&$S^{(1)}$&\textbf{0.9606}&\textbf{0.9792}&0.8051&\textbf{0.9345}\\
%			&$S^{(2)}$&0.9511&0.9742&0.7516&0.9093\\
%			&$S^{(3)}$&0.9535&0.9756&0.7490&0.9055\\
%			&SELO&0.9599&0.9788&\textbf{0.8107}&0.9187\\
%			\hline
%			\multirow{6}{*}{Bitcoin-OTC}&$A$&0.9398&0.9671&0.8070&0.9243\\
%			&$W$&0.9395&0.9667&0.8214&0.9379\\
%			&$S^{(1)}$&0.9547&0.9751&0.8622&0.9552\\
%			&$S^{(2)}$&0.9443&0.9695&0.8276&0.9341\\
%			&$S^{(3)}$&0.9445&0.9696&0.8231&0.9304\\
%			&SELO&\textbf{0.9553}&\textbf{0.9754}&\textbf{0.8656}&\textbf{0.9532}\\
%			\bottomrule
%		\end{tabular}
%	\end{center}
%\end{table}
%%%%%%%%%%%%%%%%%%%%%%%%%%%%%%%%%%%%%%%%%%%%%%%%%%%%%%%%%%%%%%%%%%%%%%%%%%%%%%%

\subsection{Ablation Study}

In this subsection, we conduct a series of ablation studies to further illustrate the effect of different modules in our proposed SELO. Firstly, we show the importance of encoding subgraph patterns with the three concatenated likelihood matrices $S^{(1)},S^{(2)},$ and $S^{(3)}$. To this end, we keep the entire architecture unchanged and replace the concatenated likelihood matrices feature by the adjacency matrix $A$, the weight matrix $W$, and each single likelihood matrix $S^{(k)}(k=1,2,3)$, respectively.

%%%%%%%%%%%%%%%%%%%%%%%%%%%%%%%%%%%%%%%%%%%%%%%%%%%%%%%%%%%%%%%%%%%%%%%%%%%%%%
\begin{table*}[tbp]
	\begin{center}
		\centering
		\caption{Comparison of the link sign prediction performances with different subgraph encoding methods. The subgraph feature is encoded by its adjacency matrix $A$,  weight matrix $W$, each single similarity matrix $S^{(1)}$, $S^{(1)}$, $S^{(1)}$ and their concatenated similarity matrices in SELO.}
		\label{tab:single_model}
		\resizebox{0.8\textwidth}{!}{
			\begin{tabular}{cc|cccccc}
				\toprule
				Dataset&Metrics&$A$&$W$&$S^{(1)}$&$S^{(2)}$&$S^{(3)}$&SELO\\
				\midrule
				\multirow{4}{*}{Bitcoin-Alpha}&Micro-F1&0.9479&0.9466&\textbf{0.9606}&0.9511&0.9535&0.9599\\
				&F1&0.9726&0.9717&\textbf{0.9792}&0.9742&0.9756&0.9788\\
				&Macro-F1&0.7260&0.7424&0.8051&0.7516&0.7490&\textbf{0.8107}\\
				&AUC&0.8823&0.9170&\textbf{0.9345}&0.9093&0.9055&0.9187\\
				\midrule
				\multirow{4}{*}{Bitcoin-OTC}&Micro-F1&0.9398&0.9395&0.9547&0.9443&0.9445&\textbf{0.9553}\\
				&F1&0.9671&0.9667&0.9751&0.9695&0.9696&\textbf{0.9754}\\
				&Macro-F1&0.8070&0.8214&0.8622&0.8276&0.8231&\textbf{0.8656}\\
				&AUC&0.9243&0.9379&\textbf{0.9552}&0.9341&0.9304&0.9532\\
				\midrule
				\multirow{4}{*}{WikiElec}&Micro-F1&0.8358&0.8135&0.8751&0.8363&0.7996&\textbf{0.8756}\\
				&F1&0.9004&0.8864&0.9222&0.9007&0.8816&\textbf{0.9225}\\
				&Macro-F1&0.7165&0.6820&\textbf{0.8076}&0.7164&0.6133&0.8031\\
				&AUC&0.8275&0.8284&\textbf{0.9148}&0.8480&0.7823&0.9138\\
				\midrule
				\multirow{4}{*}{WikiRFA}&Micro-F1&0.8188&0.8077&0.8664&0.8275&0.8011&\textbf{0.8673}\\
				&F1&0.8907&0.8835&0.9165&0.8958&0.8821&\textbf{0.9173}\\
				&Macro-F1&0.6796&0.6662&0.7901&0.6971&0.6229&\textbf{0.7908}\\
				&AUC&0.8054&0.8142&0.9044&0.8388&0.7923&\textbf{0.9052}\\
				\bottomrule
		\end{tabular}}
	\end{center}
\end{table*}
%%%%%%%%%%%%%%%%%%%%%%%%%%%%%%%%%%%%%%%%%%%%%%%%%%%%%%%%%%%%%%%%%%%%%%%%%%%%%%

The comparison results are presented in Table \ref{tab:single_model}. It can be observed that: 
\begin{itemize}
	\item Encoding the subgraph by the adjacency matrix $A$ results a worse performance. Encoding the subgraph by the weight matrix $W$ improves performance compared to that by the adjacency matrix $A$, which indicates that the hierarchical structure information and importance of negative links encoded by the re-weighting strategy is not ignorable for link sign prediction.
	\item The prediction performances are significantly improved by encoding the subgraph by the likelihood matrices instead of the adjacency matrix or weight matrix. This observation confirms that the likelihood matrix gives a better characterization of the structural features in the subgraph for determination of the target link sign.
	\item  Encoding the subgraph by one of the three likelihood matrices $S^{(1)}, S^{(2)},$ and $S^{(3)}$ can also provide acceptable prediction performances. However, the prediction metrics by the three concatenated likelihood matrices gives a better-balanced results than those by each single matrix, meaning that the three likelihood matrices contains complementary subgraph information.
\end{itemize}

\begin{figure*}[!t]
	\centering
	\subfigure[Bitcoin-Alpha]{
		\includegraphics[width=0.45\linewidth]{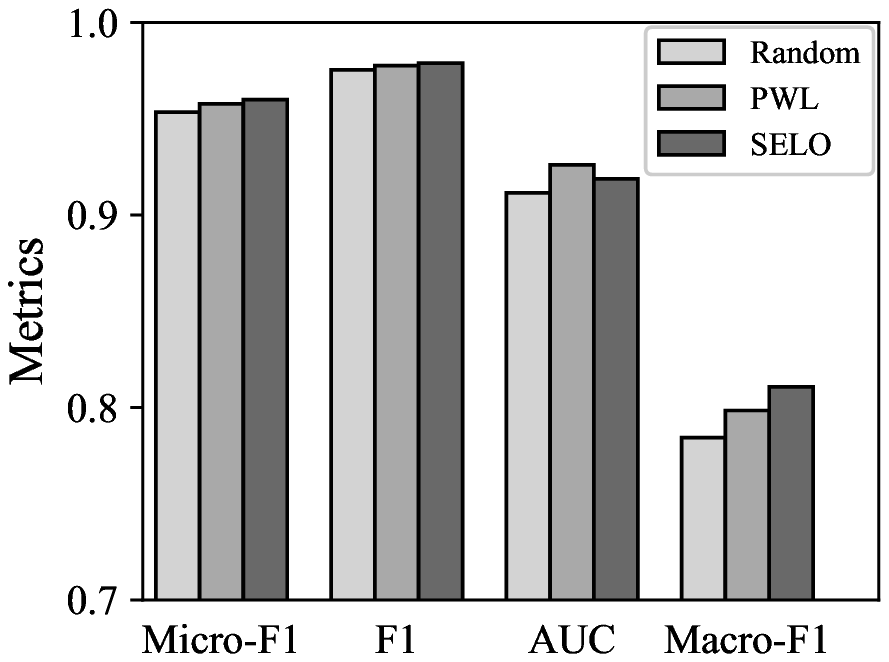}
	}
	\subfigure[Bitcoin-OTC]{
		\includegraphics[width=0.45\linewidth]{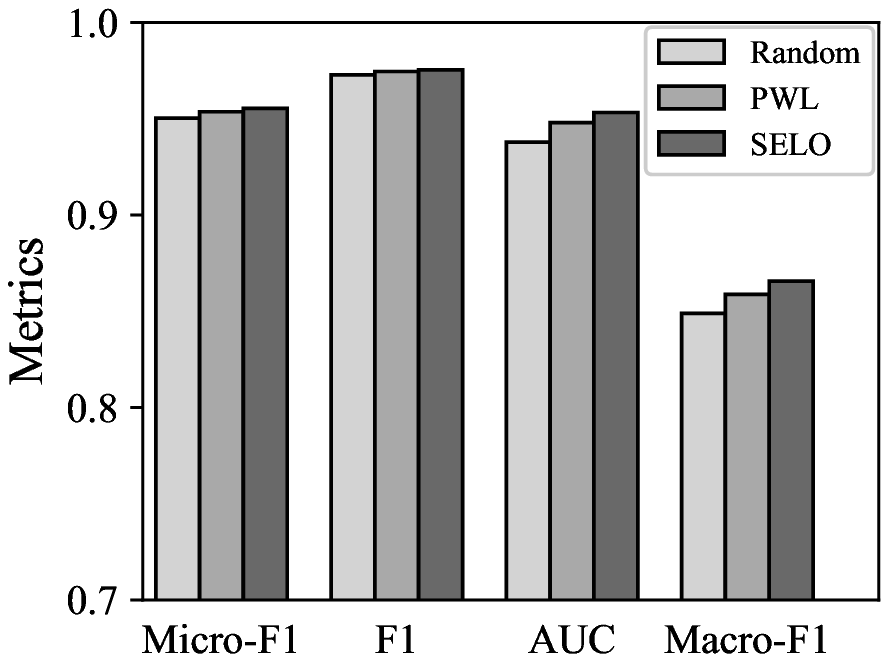}
	}
	\caption{Comparison of the link sign prediction performance with different vertex ordering methods. Random refers to that the vertices in the subgraph are labeled in a random order. PWL is a graph labeling process proposed in \cite{WLNM}.}
	\label{fig_c}
\end{figure*}

%%%%%%%%%%%%%%%%%%%%%%%%%%%%%%%%%%%%%%%%%%%%%%%%%%%%%%%%%%%%%%%%%%%%%%%%%%%%%%%
%\begin{table}[tbp]
%	\begin{center}
%		\renewcommand{\arraystretch}{1.2}
%		\centering
%		\caption{Comparison of the link sign prediction performance with different vertex ordering methods. Random refers to that the vertices in the subgraph are labeled in a random order. PWL is a graph labeling process proposed in \cite{WLNM}. }
%		\label{tab:ablation_study}
%		%\resizebox{\textwidth}{!}{
%		\begin{tabular}{cc|cccc}
%			\toprule
%			Dataset&Method&Micro-F1&F1&Macro-F1&AUC\\
%			\midrule
%			\multirow{3}{*}{Bitcoin-Alpha}
%			&Random&0.9534&0.9753&0.7843&0.9114\\
%			&PWL&0.9577&0.9776&0.7984&\textbf{0.9260}\\
%			&SELO&\textbf{0.9599}&\textbf{0.9788}&\textbf{0.8107}&0.9187\\
%			\midrule
%			\multirow{3}{*}{Bitcoin-OTC}
%			&Random&0.9503&0.9727&0.8489&0.9378\\
%			&PWL&0.9536&0.9745&0.8587&0.9483\\
%			&SELO&\textbf{0.9553}&\textbf{0.9754}&\textbf{0.8656}&\textbf{0.9532}\\
%			\bottomrule
%		\end{tabular}%}
%	\end{center}
%\end{table}
%%%%%%%%%%%%%%%%%%%%%%%%%%%%%%%%%%%%%%%%%%%%%%%%%%%%%%%%%%%%%%%%%%%%%%%%%%%%%%%

To illustrate the effect of node ordering method, we further conduct experiments with a random ordering method and the PWL (Palette Weisfeiler-Lehman) method for comparison. The random ordering method refers to that the nodes except the target pair of nodes in the subgraph are labeled in a random order. The PWL method is a graph labeling process proposed in \cite{WLNM}. Both these two methods ignore the sign of links. The comparison results are shown in Fig. \ref{fig_c}. It can be observed that the proposed node ordering method (\ref{eq:importance score}) in this paper generally improves the performance of link sign prediction.

\section{Conclusion} \label{conclusion}
%In this paper, we propose a novel SELO model to learn subgrpah topological features for link sign prediction. We first extract a enclosing subgraph for target links. Second, we weight links in subgraph by two weighting strategies. Third, we propose three linear optimization models to obtain the likehood matrices and the importance scores caculated from that matrices decide vertics ordering. Finally, the likehood matrices is fed in neural network to predict the signs on the links. We evaluate SELO model in six  real-world signed networks. The results demonstrate that SELO achieves the state-of-the-art performance, evincing that the likehood matrices mining the importance of diffirent relations. In addition, the linear models reveal that 3-hop paths are important to link sign prediction. In futurework, we plan to generalize the SELO model to weighted network or heterogeneous networks.

In this paper, we have proposed a novel SELO architecture to learn signed subgrpah topological features for link sign prediction. In detail, to predict the sign (either positive or negative) of a given target link, the proposed SELO proceeds as follows: extracting a subgraph of the target link, re-weighting the edges in the subgraph based on their signs and distances to the target link, transforming the obtained weight matrix into a similarity matrix based on the linear optimization model, pruning the similarity matrix into fixed size through vertices ordering and relabeling, and finally feeding the fixed-size similarity matrix into a fully connected neural network. 

We have evaluated the SELO architecture in five real-world signed networks by Micro-F1, F1, Macro-F1, and AUC. The experiments demonstrated that SELO achieved the best prediction performance, as compared to those baseline methods, including ALL23, SiNE, SGCN, SiGAT, SDGNN, etc. The results in this paper partially indicates the fact that the likelihood matrix obtained by the linear optimization method is more efficient than the adjacency matrix to encode the structural relevance in signed graphs. A future work is how to generalize the proposed SELO architecture to attributed signed graphs.

\bibliographystyle{IEEEtran}
\bibliography{sign_prediction_cite}

\begin{thebibliography}{10}
\providecommand{\url}[1]{#1}
\csname url@samestyle\endcsname
\providecommand{\newblock}{\relax}
\providecommand{\bibinfo}[2]{#2}
\providecommand{\BIBentrySTDinterwordspacing}{\spaceskip=0pt\relax}
\providecommand{\BIBentryALTinterwordstretchfactor}{4}
\providecommand{\BIBentryALTinterwordspacing}{\spaceskip=\fontdimen2\font plus
\BIBentryALTinterwordstretchfactor\fontdimen3\font minus
  \fontdimen4\font\relax}
\providecommand{\BIBforeignlanguage}[2]{{%
\expandafter\ifx\csname l@#1\endcsname\relax
\typeout{** WARNING: IEEEtran.bst: No hyphenation pattern has been}%
\typeout{** loaded for the language `#1'. Using the pattern for}%
\typeout{** the default language instead.}%
\else
\language=\csname l@#1\endcsname
\fi
#2}}
\providecommand{\BIBdecl}{\relax}
\BIBdecl

\bibitem{survey}
L.~L{\"u} and T.~Zhou, ``Link prediction in complex networks: A survey,''
  \emph{Physica A: statistical mechanics and its applications}, vol. 390,
  no.~6, pp. 1150--1170, 2011.

\bibitem{SEAL}
M.~Zhang and Y.~Chen, ``Link prediction based on graph neural networks,'' in
  \emph{Advances in Neural Information Processing Systems}, 2018, pp.
  5165--5175.

\bibitem{SHFF}
Z.~Liu, D.~Lai, C.~Li, and M.~Wang, ``Feature fusion based subgraph
  classification for link prediction,'' in \emph{Proceedings of the 29th ACM
  International Conference on Information \& Knowledge Management}, 2020, pp.
  985--994.

\bibitem{9174790}
Z.~Wang, Y.~Lei, and W.~Li, ``Neighborhood attention networks with adversarial
  learning for link prediction,'' \emph{IEEE Transactions on Neural Networks
  and Learning Systems}, vol.~32, no.~8, pp. 3653--3663, 2021.

\bibitem{9046288}
Z.~Wu, S.~Pan, F.~Chen, G.~Long, C.~Zhang, and P.~S. Yu, ``A comprehensive
  survey on graph neural networks,'' \emph{IEEE Transactions on Neural Networks
  and Learning Systems}, vol.~32, no.~1, pp. 4--24, 2021.

\bibitem{kumar2018community}
S.~Kumar, W.~L. Hamilton, J.~Leskovec, and D.~Jurafsky, ``Community interaction
  and conflict on the web,'' in \emph{Proceedings of the 2018 world wide web
  conference}, 2018, pp. 933--943.

\bibitem{leskovec2010predicting}
J.~Leskovec, D.~Huttenlocher, and J.~Kleinberg, ``Predicting positive and
  negative links in online social networks,'' in \emph{Proceedings of the 19th
  international conference on World wide web}, 2010, pp. 641--650.

\bibitem{chiang2011exploiting}
K.-Y. Chiang, N.~Natarajan, A.~Tewari, and I.~S. Dhillon, ``Exploiting longer
  cycles for link prediction in signed networks,'' in \emph{Proceedings of the
  20th ACM international conference on Information and knowledge management},
  2011, pp. 1157--1162.

\bibitem{beigi2020social}
G.~Beigi, J.~Tang, and H.~Liu, ``Social science--guided feature engineering: A
  novel approach to signed link analysis,'' \emph{ACM Transactions on
  Intelligent Systems and Technology (TIST)}, vol.~11, no.~1, pp. 1--27, 2020.

\bibitem{BESIDE}
Y.~Chen, T.~Qian, H.~Liu, and K.~Sun, ````bridge'' enhanced signed directed
  network embedding,'' in \emph{Proceedings of the 27th ACM International
  Conference on Information and Knowledge Management}, 2018, pp. 773--782.

\bibitem{SGCN}
T.~Derr, Y.~Ma, and J.~Tang, ``Signed graph convolutional networks,'' in
  \emph{2018 IEEE International Conference on Data Mining (ICDM)}.\hskip 1em
  plus 0.5em minus 0.4em\relax IEEE, 2018, pp. 929--934.

\bibitem{SNEA}
Y.~Li, Y.~Tian, J.~Zhang, and Y.~Chang, ``Learning signed network embedding via
  graph attention,'' in \emph{Proceedings of the AAAI Conference on Artificial
  Intelligence}, vol.~34, no.~04, 2020, pp. 4772--4779.

\bibitem{SDGNN}
J.~Huang, H.~Shen, L.~Hou, and X.~Cheng, ``Sdgnn: Learning node representation
  for signed directed networks,'' in \emph{Proceedings of the AAAI Conference
  on Artificial Intelligence}, vol.~35, no.~1, 2021, pp. 196--203.

\bibitem{GS-GNN}
H.~Liu, Z.~Zhang, P.~Cui, Y.~Zhang, Q.~Cui, J.~Liu, and W.~Zhu, ``Signed graph
  neural network with latent groups,'' in \emph{Proceedings of the 27th ACM
  SIGKDD international conference on Knowledge discovery and data mining},
  2021, pp. 1066--1075.

\bibitem{WLNM}
M.~Zhang and Y.~Chen, ``Weisfeiler-lehman neural machine for link prediction,''
  in \emph{Proceedings of the 23rd ACM SIGKDD International Conference on
  Knowledge Discovery and Data Mining}, 2017, pp. 575--583.

\bibitem{LO}
R.~Pech, D.~Hao, Y.-L. Lee, Y.~Yuan, and T.~Zhou, ``Link prediction via linear
  optimization,'' \emph{Physica A: Statistical Mechanics and its Applications},
  vol. 528, p. 121319, 2019.

\bibitem{cho2006mechanism}
J.~Cho, ``The mechanism of trust and distrust formation and their relational
  outcomes,'' \emph{Journal of retailing}, vol.~82, no.~1, pp. 25--35, 2006.

\bibitem{tang2016survey}
J.~Tang, Y.~Chang, C.~Aggarwal, and H.~Liu, ``A survey of signed network mining
  in social media,'' \emph{ACM Computing Surveys (CSUR)}, vol.~49, no.~3, pp.
  1--37, 2016.

\bibitem{tang2014distrust}
J.~Tang, X.~Hu, and H.~Liu, ``Is distrust the negation of trust? the value of
  distrust in social media,'' in \emph{Proceedings of the 25th ACM conference
  on Hypertext and social media}, 2014, pp. 148--157.

\bibitem{ma2009learning}
H.~Ma, M.~R. Lyu, and I.~King, ``Learning to recommend with trust and distrust
  relationships,'' in \emph{Proceedings of the third ACM conference on
  Recommender systems}, 2009, pp. 189--196.

\bibitem{guha2004propagation}
R.~Guha, R.~Kumar, P.~Raghavan, and A.~Tomkins, ``Propagation of trust and
  distrust,'' in \emph{Proceedings of the 13th international conference on
  World Wide Web}, 2004, pp. 403--412.

\bibitem{Devi}
S.~J. Devi and B.~Singh, ``Link prediction model based on the topological
  feature learning for complex networks,'' \emph{ARABIAN JOURNAL FOR SCIENCE
  AND ENGINEERING}, vol.~45, no.~12, pp. 10\,051--10\,065, 2020.

\bibitem{WL}
B.~Weisfeiler and A.~A. Lehman, ``A reduction of a graph to a canonical form
  and an algebra arising during this reduction,'' \emph{Nauchno-Technicheskaya
  Informatsia}, vol.~2, no.~9, pp. 12--16, 1968.

\bibitem{deepwalk}
B.~Perozzi, R.~Al-Rfou, and S.~Skiena, ``Deepwalk: Online learning of social
  representations,'' in \emph{Proceedings of the 20th ACM SIGKDD international
  conference on Knowledge discovery and data mining}, 2014, pp. 701--710.

\bibitem{node2vec}
A.~Grover and J.~Leskovec, ``node2vec: Scalable feature learning for
  networks,'' in \emph{Proceedings of the 22nd ACM SIGKDD international
  conference on Knowledge discovery and data mining}, 2016, pp. 855--864.

\bibitem{LINE}
J.~Tang, M.~Qu, M.~Wang, M.~Zhang, J.~Yan, and Q.~Mei, ``Line: Large-scale
  information network embedding,'' in \emph{Proceedings of the 24th
  international conference on world wide web}, 2015, pp. 1067--1077.

\bibitem{SiNE}
S.~Wang, J.~Tang, C.~Aggarwal, Y.~Chang, and H.~Liu, ``Signed network embedding
  in social media,'' in \emph{Proceedings of the 2017 SIAM international
  conference on data mining}.\hskip 1em plus 0.5em minus 0.4em\relax SIAM,
  2017, pp. 327--335.

\bibitem{SIGNet}
M.~R. Islam, B.~A. Prakash, and N.~Ramakrishnan, ``Signet: Scalable embeddings
  for signed networks,'' in \emph{Pacific-Asia Conference on Knowledge
  Discovery and Data Mining}.\hskip 1em plus 0.5em minus 0.4em\relax Springer,
  2018, pp. 157--169.

\bibitem{SiGAT}
J.~Huang, H.~Shen, L.~Hou, and X.~Cheng, ``Signed graph attention networks,''
  in \emph{International Conference on Artificial Neural Networks}.\hskip 1em
  plus 0.5em minus 0.4em\relax Springer, 2019, pp. 566--577.

\bibitem{adam}
Y.~Bengio, J.~Louradour, R.~Collobert, and J.~Weston, ``Curriculum learning,''
  in \emph{Proceedings of the 26th annual international conference on machine
  learning}, 2009, pp. 41--48.

\end{thebibliography}

\end{document}